\documentclass[conference]{IEEEtran}
\usepackage{amsmath,amssymb,amsfonts}
\usepackage{algorithmic}
\usepackage{algorithm}
\usepackage{graphicx}
\usepackage{textcomp}
\usepackage{braket}
\usepackage{xcolor}
\usepackage{url}
\usepackage[hidelinks]{hyperref} % Optional: for clickable links
\usepackage{booktabs} % For potential tables later
\usepackage{subcaption} % For subfigures
\usepackage{soul}

\usepackage{amsthm}
\usepackage[nameinlink,noabbrev]{cleveref}
\theoremstyle{definition}
\newtheorem{definition}{Definition}

\begin{document}

\title{Enhancing Federated Learning Privacy with QUBO}
% \author{Anonymous Authors}

\author{\IEEEauthorblockN{1\textsuperscript{st} Andras Ferenczi}
\IEEEauthorblockA{
\textit{American Express Co}\\andras.l.ferenczi1@aexp.com}
\and
\IEEEauthorblockN{2\textsuperscript{nd} Sutapa Samanta}
\IEEEauthorblockA{
\textit{American Express Co}\\sutapasamanta07@gmail.com}
\and
\IEEEauthorblockN{3\textsuperscript{rd} Dagen Wang}
\IEEEauthorblockA{
\textit{American Express Co}\\Dagen.Wang@aexp.com}
\and
\IEEEauthorblockN{4\textsuperscript{th} Todd Hodges}
\IEEEauthorblockA{
\textit{American Express Co}\\Todd.Hodges@aexp.com}
}

\maketitle

\begin{abstract}
Federated learning (FL) is a widely used method for training machine learning (ML) models in a scalable way while preserving privacy (i.e., without centralizing raw data). Prior research shows that the risk of exposing sensitive data increases cumulatively as the number of iterations where a client's updates are included in the aggregated model increase. Attackers can launch membership inference attacks (MIA; deciding whether a sample or client participated), property inference attacks (PIA; inferring attributes of a client’s data), and model inversion attacks (MI; reconstructing inputs), thereby inferring client-specific attributes and, in some cases, reconstructing inputs. In this paper, we mitigate risk by substantially reducing per client exposure using a quantum computing–inspired quadratic unconstrained binary optimization (QUBO) formulation that selects a small subset of client updates most relevant for each training round. In this work, we focus on two threat vectors: (i) information leakage by clients during training and (ii) adversaries who can query or obtain the global model. We assume a trusted central server and do not model server compromise. This method also assumes that the server has access to a validation/test set with global data distribution. Experiments on the MNIST dataset with 300 clients in 20 rounds showed a 95.2\% per-round and 49\% cumulative privacy exposure reduction, with 147 clients' updates never being used during training while maintaining in general the full-aggregation accuracy or even better. The method proved to be efficient at lower scale and more complex model as well. A CINIC-10 dataset-based experiment with 30 clients resulted in 82\% per-round privacy improvement and 33\% cumulative privacy.
\end{abstract}


\section{Introduction and Motivation}
Federated learning (FL) is a method of training a global model by aggregating local updates and re-distributing an aggregated model across multiple rounds \cite{mcmahan2017communication}. FL is not immune to common privacy risks, including gradient inversion attacks (GIA)~\cite{zhu2019deep}, membership inference attacks (MIA) \cite{shokri2017membership}, property inference attacks (PIA) \cite{melis2019exploiting}, and model inversion attacks (MI)~\cite{fredrikson2015model}, which may result in leakage of sensitive data.

Access to per-round global model snapshots enables multi-snapshot membership inference that extracts signals unavailable from a single final model, thereby amplifying leakage as more snapshots accumulate~\cite{chang2024efficient,suri2022subject}. The scenarios in scope of this study include an honest server that does the aggregation, with potential semi-honest and curious clients (FL participants) that have access to the global model snapshot after each round. Slope-based auditing shows that tracking the logit and loss for a particular example of its true class results in a steeper slope than for non-member samples \cite{chang2024efficient}. Subject-level black-box inference attacks are effective on the final global model but even more so if the attacker has access to snapshots after each round; Suri et al. conclude that the efficacy of such attacks increases with the number of rounds~\cite{suri2022subject}. Clients included more frequently have a higher influence on the aggregated model and hence are more vulnerable. In general, theory shows that combining information from successive releases strengthens membership inference~\cite{jagielski2023combine}. Paper~\cite{wu2024fedinverse} proposes FedInverse, a method that reconstructs training data in FL by placing the attacker in the participant role and using Generative Adversarial Network (GAN)-based model. Authors of~\cite{so2023securingsecureaggregationmitigating}, also point out that exposure grows with participation frequency.

We propose a novel approach that leverages Quadratic Unconstrained Binary Optimization (QUBO)~\cite{glover2018tutorial} to reduce participation frequency in FL through strategic client selection. The key insight in this proposal is that client selection can be optimized in such manner as not to impact the quality of the training, thus reducing exposure to privacy attacks by men-in-the-middle or curious training participants, assuming trusted server scenarios. As QUBO formulations are particularly well-suited for quantum annealers, the proposal offers potential at scale as quantum hardware matures.

Our contributions are threefold: (1) We formulate federated learning client selection as a QUBO optimization problem that balances client relevance, diversity, and redundancy; (2) We introduce ten distinct QUBO strategies spanning the exploration-exploitation spectrum, from consensus-focused approaches to diversity-maximizing selections; (3) We provide comprehensive experimental evaluation on MNIST~\cite{lecun1998mnist} with 300 clients and CINIC-10~\cite{darlow2018cinic10} with 30 clients across varying data heterogeneity settings using Dirichlet distribution \cite{li2020federated}, demonstrating client selection efficiency while maintaining similar accuracy to the standard aggregated global model.

The remainder of this paper is organized as follows: Section~\ref{sec:background} provides background on FL, surveys related work in FL client selection, motivation for choosing optimization-based client selection, and introduction to QUBO. Section~\ref{sec:proposal} details our proposed QUBO-based client selection approach. Experimental methodology and results were discussed in Section~\ref{sec:experiments}. Section~\ref{sec:discussion} discusses implications and limitations of our experiments and conclude with future research directions.

\section{Background}\label{sec:background}
\subsection{Federated Learning (FL)}

FL is a method used to train a common ``global" model by aggregating gradient updates from distributed clients that train their copies of the model using local data \cite{mcmahan2017communication}. A central server maintains global parameters $\mathbf{w}^t$ at round $t$. Client $i$ receives the global model, performs local training on its data $\mathcal{D}_i$, and computes parameter updates $\Delta \mathbf{w}_i^t = \mathbf{w}_i^{t+1} - \mathbf{w}^t$. The server aggregates client updates using weighted averaging:
\begin{equation*}
\mathbf{w}^{t+1} = \mathbf{w}^t + \sum_{i \in \mathcal{S}_t} \frac{|\mathcal{D}_i|}{|\mathcal{D}|} \Delta \mathbf{w}_i^t,
\end{equation*}
where $\mathcal{S}_t$ is the set of selected clients in round $t$, $|\mathcal{D}_i|$ is the number of samples at client $i$, and $|\mathcal{D}| = \sum_{i \in \mathcal{S}_t} |\mathcal{D}_i|$. Local training typically minimizes the loss function
\begin{equation*}
\mathcal{L}_i(\mathbf{w}) = \frac{1}{|\mathcal{D}_i|} \sum_{(\mathbf{x}, y) \in \mathcal{D}_i} \ell(\mathbf{w}; \mathbf{x}, y),
\end{equation*}
where $\ell$ is the sample-wise loss function.

\subsection{Efficient Client Selection in FL}
Extensive literature exists on methods for efficient client selection, most of which aim to speed up training and reduce communication rounds, minimize network bandwidth usage, or select trustworthy updates in Byzantine settings.

The original \emph{FedAvg} paper~\cite{mcmahan2017communication} proposes random selection of clients to reduce processing and network overhead. In contrast, Power-of-choice client selection \cite{cho2020client} describes a method in which a central server pools a random set of clients for their training loss and selects a subset with the highest loss, favoring clients with the highest loss (i.e., ``harder" clients) to speed up convergence. The problem with this method is that it fails to ensure convergence, especially when training data is heterogeneous across clients, in which case the authors reduce the loss-based selection criteria.

\emph{FedNova}~\cite{wang2020tackling} is pointing out that in ``regular" FL the server computes FedAvg irrespective of the number of epochs performed by each client, thus using the incorrect objective function. The paper proposes that each client sends a normalized update, which in simplest case, consists of the sum of gradients over multiple epochs divided by the number of epochs. Even though the authors don't provide customization based on statistical heterogeneity, they point out that their solution works as it optimizes the correct objective function.

\emph{Oort}~\cite{lai2021oort} also uses self-reported loss based on the pre-defined loss function, plus benchmarks of client network communications efficiency to define the statistical utility of a client: 
\begin{equation}\label{eq:statistical_utility}
    U(i) = |\mathcal{D}_i|\,\sqrt{\frac{1}{|\mathcal{D}_i|}\sum_{k \in \mathcal{D}_i}\bigl(\mathrm{Loss}(k)\bigr)^2},
\end{equation}
where $i$ indexes clients, $\mathcal{D}_i$ is client $i$'s dataset,  $|\mathcal{D}_i|$ is dataset size, $k \in \mathcal{D}_i$ is a sample, and $\mathrm{Loss}(k)$ the training loss on $k$. 
\[
\mathrm{Util}(i) = U(i)\,\left(\frac{T}{t_i}\right)^{\mathbf{1}(T < t_i)\,\alpha},
\]
where $\mathrm{Util}(i)$ is the selection utility, $ U(i)$ is the statistical utility as defined in Eqn.~\eqref{eq:statistical_utility}, $t_i$ is the estimated round time for client $i$, $T$ is the preferred round duration, $\mathbf{1}(x)$ is an indicator function that takes value 1 if $x$ is true and 0 otherwise, $\mathbf{1}(T<t_i)$ is the indicator that $t_i\!>\!T$, and $\alpha$ is the penalty exponent. The method assumes honest self-reporting and knowledge of the client's data characteristics.

\emph{FedCS}~\cite{Nishio_2019} describes a meta protocol for choosing the most optimal mobile clients to fit the aggregation within a given timeframe. The characteristics of mobile devices is collected by a mobile edge computing operator. In order for an edge device to be selected, the following inequality must stand:
% Minimal core constraint (essential)
\[
T_{\mathrm{round}} \;\ge\; T_S \;+\; \frac{D_m}{\min_{k\in S}\theta^{\mathrm{DL}}_k}
\;+\; \Theta_{|S|} \;+\; T_{\mathrm{Agg}} .
\]
% Symbols (brief, each with a parenthetical explanation):
Here $T_{\mathrm{round}}$ is the total time budget for one federated round, $T_S$ is the selection overhead, i.e., time to poll clients and compute the schedule,  $D_m$ is the model size, i.e., number of bits to distribute, $\theta^{\mathrm{DL}}_k$ is the downlink throughput for client $k$ measured in bits/s. The quantity $\min_{k\in S}\theta^{\mathrm{DL}}_k$ represents the slowest downlink among selected clients.
$\Theta_{|S|}$ is the cumulative update plus upload time elapsed until the last selected client finishes, accounting for overlap.
$T_{\mathrm{Agg}}$ is the server aggregation time, i.e., the time to average updates. 

Post-update ``selection" via robust aggregation further preserves model integrity, e.g., \emph{Krum}/\emph{Multi-Krum} pick a few mutually consistent updates, \emph{Bulyan} combines \emph{Krum} with coordinate-wise trimming, and coordinate-wise \emph{Trimmed Mean}/median downweight outliers \cite{blanchard2017krum,guerraoui2018bulyan,yin2018trimmedmean}.

QUBO-optimization has been proposed to address different FL challenges. For example, \cite{10636852} presents a hybrid quantum–classical algorithm to optimize how FL models share computing resources across a distributed network.

\subsection{Motivation for Optimization-Based Client Selection}

While existing FL client selection methods offer valuable insights, they predominantly rely on heuristic approaches that fail to capture the fundamental combinatorial nature of the selection problem. Client selection in FL inherently involves choosing a subset of clients from a larger pool -- a classic combinatorial optimization challenge that requires balancing multiple competing objectives simultaneously.

Current methods typically optimize single metrics, e.g., loss values in Power-of-choice, utility scores in \emph{Oort}, or apply simple aggregation rules. However, an effective client selection must consider complex trade-offs between client relevance, i.e., how much each client contributes to model improvement, and redundancy which avoids selection of clients with similar data distributions. These competing objectives naturally lead to quadratic interactions in the optimization landscape, where pairwise relationships between clients become as important as individual client characteristics.

Furthermore, privacy-conscious FL systems aim to minimize the number of clients whose sensitive gradient information is exposed to the central server. This introduces an additional constraint of selecting the smallest effective subset of clients while maintaining model quality. Traditional heuristic methods lack principled frameworks for incorporating such privacy considerations into the selection process.

These limitations motivate the need for a more systematic optimization approach that can simultaneously handle multiple objectives, capture pairwise client relationships, and incorporate practical constraints such as privacy requirements and communication efficiency.

\subsection{Quadratic Unconstrained Binary Optimization}

QUBO is a mathematical optimization framework where binary decision variables are optimized over quadratic objective functions~\cite{lucas2014ising}. The idea is to minimize an objective function of the form
\begin{equation*}
\min_{\mathbf{x} \in \{0,1\}^n} \mathbf{x}^T Q \mathbf{x},
\end{equation*}
where $Q$ is an $n \times n$ real-valued matrix and $\mathbf{x}$ is a binary decision vector. QUBO problems can be efficiently solved using quantum annealers or simulated annealing approaches. Although popularized through quantum annealing, QUBO has long been studied in classical optimization and has proven effective for problems involving binary decision variables. The framework's key strength lies in its ability to naturally accommodate competing objectives through quadratic terms, making it particularly well-suited for combinatorial problems with complex trade-offs.

For FL client selection, QUBO provides several critical advantages. First, the binary nature of the decision variables $x_i \in \{0,1\}$ representing whether client $i$ is selected, directly matches the selection problem structure. Second, QUBO's quadratic formulation enables sophisticated modeling of pairwise relationships between clients, essential for capturing redundancy between clients with similar data distributions. Third, modern QUBO solvers, including quantum annealers and classical simulated annealing algorithms, can efficiently explore large solution spaces that would be computationally prohibitive for brute-force enumeration.

The quadratic objective function allows simultaneous optimization of multiple goals: linear terms can encode individual client relevance scores, while quadratic terms capture interactions such as redundancy penalties and constraint enforcement. This unified framework eliminates the need for complex multi-stage heuristics and provides principled mechanisms for incorporating privacy constraints and communication efficiency requirements.

To the best of our knowledge, QUBO has not been widely applied to optimize FL aggregation, though it has proven successful in other ML applications such as support-vector machines, linear regression, and balanced $k$-means clustering. QUBO is frequently used in various business domains, including finance for portfolio optimization and asset allocation, biology and drug discovery, and logistics and transportation. To best of our knowledge, this work represents one of the first applications of QUBO to FL client selection with explicit privacy considerations.

\section{Proposed QUBO-Based Client Selection}\label{sec:proposal}

\begin{definition}[Privacy-preservation proxy metrics]\label{def:privacy-proxy}
We evaluate privacy via participation-rate proxies aligned with our threat model.
\begin{enumerate}
  \item \textit{Per-round privacy preservation:} $1 - \tfrac{|S_t|}{n}$,
  where \( |S_t| \) is the number of selected clients in round \(t\) and \( n \) is the client pool size.
  \item \textit{Cumulative privacy preservation:} (a) fraction of clients never selected across all rounds; and/or (b) $1 - \overline{p}$,
  where \( \overline{p} \) is the mean participation rate per client.
\end{enumerate}
These metrics reflect reduced exposure of client updates to other clients via released snapshots.
\end{definition}

\subsection{Mathematical Formulation Details}
We build a QUBO that maximizes client relevance ($r$) while minimizing the redundancy by penalizing choice of client with similar updates. For clarity in the QUBO formulation, we drop time superscripts and use $\Delta \mathbf{w}_i$ to denote client $i$'s parameter update in the current round. The relevance is measured by  computing the proximity of an update to the mean update or consensus as follows:
\begin{equation}\label{eq:relevance}
r_i = 1 - \frac{\|\Delta \mathbf{w}_i - \overline{\Delta \mathbf{w}}\|_2}{\max_j \|\Delta \mathbf{w}_j - \overline{\Delta \mathbf{w}}\|_2 + \epsilon},
\end{equation}
where $\Delta \mathbf{w}_i$ is client $i$'s parameter update, $\overline{\Delta \mathbf{w}} = \frac{1}{n}\sum_{j=1}^n \Delta \mathbf{w}_j$ is the mean update across all $n$ clients, and $\epsilon = 10^{-8}$ prevents division by zero. We use normalized relevance $r_i^\text{norm}$ defined as 
\begin{equation*}
    r_i^\text{norm} = \frac{r_i - \min_j(r_j)}{\max_j(r_j) - \min_j(r_j) + \epsilon}\ ,
\end{equation*}
where $r_i$ is defined in Eqn.~\ref{eq:relevance}.
Similarity matrix is constructed by computing pairwise cosine similarity ($S_{ij}$) between normalized client updates as follows:
\begin{equation*}
S_{ij} = \frac{\Delta \mathbf{w}_i \cdot \Delta \mathbf{w}_j}{\|\Delta \mathbf{w}_i\|_2 \|\Delta \mathbf{w}_j\|_2}, \quad S_{ii} = 0\ .
\end{equation*}
The QUBO for client selection takes the following from:
\begin{multline}\label{qubo}
    Q = -\beta_r\sum_{i=1}^n r_i^{\text{norm}} x_i + \lambda_s\sum_{i=1}^n \sum_{j=1}^n S_{ij} x_i x_j\\
    + \lambda_c \left(\sum_{i=1}^n x_i -k\right)^2\ ,
\end{multline}
where $\beta_r$ is relevance scaling factor, $\lambda_s$ is strategy-specific redundancy weight, $k$ is the target data size, i.e., number of client updates to be selected, $\lambda_c$ is penalty weight. The decision binary variable takes value $x_i = 1$ if client $i$ is selected, and $x_i =0$ if it is not selected. 

We further modify QUBO given in Eqn.~\ref{qubo} during training rounds to restrict repetitive client selection via following two mechanisms.
\begin{enumerate}
    \item \textit{Max selection constraint:} A hard constraint mechanism is used to exclude clients that exceed a maximum selection threshold. Client $i$ can be excluded from the training round by ``forcing" a hard constraint of the form:
    \begin{equation}\label{eq:max-selections-exclusion}
        \text{exclude}(i) = \begin{cases}
        \text{True} & \text{if } c_i^{(t)} \geq \text{max\_selections}, \\
        \text{False} & \text{otherwise}\ .
        \end{cases}
    \end{equation}
    \item \textit{Fairness mode:} An optional fairness mode applies aggressive penalties to over-selected clients by modifying diagonal terms with a security penalty. When enabled, clients with higher selection counts receive additional negative weights in the QUBO formulation, further discouraging repeated selection and promoting balanced participation across the client pool. This complements the max selections constraint by providing soft enforcement before hard exclusion.
\end{enumerate}

\subsection{Ten QUBO Strategies} We employ ten different client selection strategies spanning the exploration-exploitation spectrum using QUBO. We vary the penalty $\lambda_s$ from low to high to go from consensus-based strategy to diversity-based strategy. We call them Max-Consensus, Ultra-Consensus, Med-Consensus, High-Consensus, Magnitude-hybrid, Balanced, Low-Diversity, High-Diversity, Ultra-Diversity, and Max-Diversity. We also modify the QUBO depending on the strategy selected as follows:
\begin{enumerate}
    \item \textit{Anti-clustering enhancement:} For Max-Consensus and Ultra-Consensus strategies we introduce anti-clustering enhancement to the similarity matrix if the element value exceeds a certain threshold $\tau$, i.e., we set 
    \begin{equation}\label{eq:anti-clustering}
       \lambda_s = \begin{cases}
         0.3 & \text{if } S_{ij} > \tau, \\
       \lambda_r^s & \text{otherwise} \ ,
        \end{cases}
    \end{equation}
    where values of $\lambda_r^s$ for MNIST experiment are listed in Table~\ref{tab:params}. The value of $\lambda_r^s$ is chosen to be small which reduce the clustering penalty for Max-Consensus and Ultra-Consensus strategies. But we avoid too much clustering by introducing a larger penalty of $0.3$ when similarity is too high. $\tau$ is chosen to be 0.98 for MNIST data, and 0.90 for CINIC-10 data.
    \item \textit{Magnitude-boost:} For the Magnitude-Hybrid strategy, relevance is enhanced with gradient magnitude information as given in Eqn.~\ref{eq:mag_boost_relevance}.
    \begin{equation}\label{eq:mag_boost_relevance}
        r_i^\text{mag} = (1-\gamma)r_i^\text{norm} + \frac{ \gamma\|\Delta \mathbf{w}_i\|_2}{\max_j \|\Delta \mathbf{w}_j\|_2 + \epsilon}\ ,
    \end{equation}
    where $0<\gamma<1$ is a magnitude boosting parameter. Setting $\gamma=0.3$ worked well for our experiments.
\end{enumerate}
We set $\beta_r = 3.0$ across all ten strategies and both the datasets although optimal values may differ by model-dataset combination. Values of $\lambda_r^s$ and $\lambda_c$ for different QUBO strategies for the MNIST dataset are listed in Table~\ref{tab:params}. Anti-clustering penalizes high similarity using Eqn.~\ref{eq:anti-clustering}.
\begin{table}[htbp]
\begin{center}
% \small
\begin{tabular}{|l|c|c|c|c|}
\hline
Strategy & $\lambda_{r}^s$ &
$\lambda_c$ &
\begin{tabular}[c]{@{}c@{}}Anti-\\ Cluster\end{tabular} &
\begin{tabular}[c]{@{}c@{}} Mag-\\Boost\end{tabular} \\
\hline
Max-Consensus     & 0.02 & 3.0 & $\checkmark$ & $\times$ \\
Ultra-Consensus   & 0.03 & 2.0 & $\checkmark$ & $\times$ \\
High-Consensus    & 0.05 & 1.0 & $\times$     & $\times$ \\
Med-Consensus     & 0.04 & 1.5 & $\times$     & $\times$ \\
Magnitude-Hybrid  & 0.10 & 1.0 & $\times$     & $\checkmark$ \\
Balanced          & 0.15 & 0.5 & $\times$     & $\times$ \\
Low-Diversity     & 0.20 & 0.4 & $\times$     & $\times$ \\
High-Diversity    & 0.25 & 0.5 & $\times$     & $\times$ \\
Ultra-Diversity   & 0.35 & 0.3 & $\times$     & $\times$ \\
Max-Diversity     & 0.40 & 0.2 & $\times$     & $\times$ \\
\hline
\end{tabular}
\end{center}
\caption{QUBO strategy parameter values for MNIST dataset}\label{tab:params}
\end{table}
The CINIC-10 penalty-weights ($\lambda_c$) were empirically optimized through extensive experimentation. We use the values, (i)  $\lambda_c \in [0.6, 0.9]$ for diversity strategies, (ii) $\lambda_c \in [1.2, 2.0]$ for consensus strategies, and (iii) $\lambda_c = 1.0$ for balanced strategy.

\subsection{Strategy Selection Algorithm}
We compare all the ten strategies for QUBO in each round of training and select the best strategy and corresponding client selection based on the best scoring. We use a composite scoring function that balances multiple objectives as given in Eqn.~\ref{eq:scoring} 
\begin{equation}\label{eq:scoring}
score_s = \eta_1 \cdot accuracy_s +\eta_2\cdot \lambda_{r}^s - \eta_3 \cdot variance_s,
\end{equation}
where different components are defined as follows:
\begin{itemize}
\item $accuracy_s$: Test accuracy obtained by temporarily aggregating selected client updates with the global model and evaluating on server validation data,
\item $variance_s$: Mean standard deviation across parameter dimensions of the selected client updates, 

\item $\eta_1, \eta_2, \eta_3$: Dataset-adaptive composite weights optimized for each dataset. The parameter values for our experiments are 
\begin{align}
(\eta_1, \eta_2, \eta_3) = \begin{cases}
(1.033, 0.01, 1.082) & \text{CINIC-10}, \\
(1.0, 0.01, 0.001) & \text{MNIST}\ .
\end{cases}
\end{align}
\end{itemize}
The full workflow is given in Algorithm~\ref{algo:client_selection}. 
\begin{algorithm}[htbp]
\caption{QUBO-based Client Selection Algorithm}\label{algo:client_selection}
\begin{algorithmic}[1]
\REQUIRE Global parameters $\mathbf{w}^t$, client updates $\{\Delta \mathbf{w}_i\}_{i=1}^n$, selection size $k$
\ENSURE Selected client set $\mathcal{S}_t$, winning strategy name
\STATE Compute normalized client relevance scores $\{r_i^\text{norm}\}_{i=1}^n$
\STATE Compute similarity matrix $S$
\STATE Apply max selections exclusion if enabled
\STATE Initialize $best\_score = -\infty$, $best\_strategy = \text{NULL}$
\FOR{each strategy $s$ in strategies}
    \STATE Compute strategy-adaptive $\lambda_c$ based on dataset and strategy type
    \STATE Apply magnitude boost if strategy requires it
    \STATE Construct QUBO matrix $Q^s$ using complete formulation:
    \IF{strategy has magnitude-boost}
        \STATE \quad $Q_{ii} = -\beta_{r} r_i^\text{mag}   + \lambda_{c} (1 - 2k)$
    \ELSE
        \STATE \quad $Q_{ii} = -\beta_{r} r_i^\text{norm}   + \lambda_{c} (1 - 2k)$
    \ENDIF
    \IF{strategy has anti-clustering}
        \STATE \quad $Q_{ij} = 2\lambda_{c} + \begin{cases} 0.3 \cdot S_{ij} & \text{if } S_{ij} > \tau \\ \lambda_{r}^s \cdot S_{ij} & \text{otherwise} \end{cases}$
    \ELSE
        \STATE \quad $Q_{ij} = 2\lambda_{c} + \lambda_{r}^s  S_{ij}$
    \ENDIF
    \STATE Solve QUBO
    \STATE Extract selected clients: $\mathcal{S}_s = \{i : x_i^* = 1\}$
    \STATE Evaluate strategy performance: $score_s = f(\mathcal{S}_s, \mathbf{w}^t)$
    \IF{$score_s > best\_score$}
        \STATE $best\_score = score_s$, $best\_strategy = s$, $\mathcal{S}_t = \mathcal{S}_s$
    \ENDIF
\ENDFOR
\RETURN $\mathcal{S}_t$, $best\_strategy$
\end{algorithmic}
\end{algorithm}

\subsection{Differential Privacy Complementarity}

The proposed method can work in combination with Differential Privacy (DP) methods. In such a scenario, the central server would apply DP on the final global model that is shared with other entities, if applicable. The assumption is that the network connectivity between the server and the clients is secured, no eavesdropping is possible, and the server is neither curious nor malicious.

% \section{Methodology: Quantum Embedding for Aggregation}\label{sec:methodology}
% \input{methodology}

\section{Experimental Plan and Results}\label{sec:experiments}
\subsection{Experimental Setup}

We conducted a series of comprehensive tests on public datasets to validate our assumptions. First, we consider MNIST handwritten digits~\cite{lecun1998mnist} with $60,000$ training, and $10,000$ test samples. The MNIST experiment involves 300 simulated clients and one server. We use Dirichlet distribution~\cite{li2020federated} of data with varying $\alpha$ parameters to control  data heterogeneity. We use three different levels of heterogeneity. They are (i) extreme heterogeneity with $\alpha = 0.001$, (ii) high heterogeneity with $\alpha = 0.01$, and (iii) moderate heterogeneity with $\alpha = 0.1$.
We use simulated annealing (SA) to solve the QUBO. The QUBO strategies were configured to select 10 clients with flexibility to select less or more for optimal result. We consider Convolutional Neural Network (CNN) with the following architecture as our ML model:
\begin{itemize}
\item Conv2D($1\rightarrow 32, 3\times 3$) $\rightarrow$ ReLU $\rightarrow$ MaxPool2D($2\times2$)
\item Conv2D($32\rightarrow64, 3\times3$) $\rightarrow$ ReLU $ \rightarrow$ MaxPool2D($2\times2$)
\item Flatten $\rightarrow$ Dense($1600 \rightarrow 128$) $\rightarrow$ ReLU $\rightarrow$ Dense(128$\rightarrow$10)
\end{itemize}
We train the  ML for 20 rounds, each round consisting of 20 iterations. We set FedAvg learning rate to be 0.065, and QUBO client selection learning rate to be 0.082.

\subsection{Privacy-Preserving Baseline Comparison}
We run three independent experiments in parallel -- (i) the standard FedAvg method that aggregates all 300 client updates; (ii) the QUBO method that is configured for 10 clients with flexibility to select more or less as needed (selected 13.95 in average); and finally (iii) the Random method that selects same number of clients as the QUBO method. Notably, all three methods use FedAvg for aggregation -- the difference lies in the selection of clients that participate in the aggregation. QUBO selection strategy uses only 13.95 out if 300 clients each round preserving 95.2\% of client's privacy. 

We analyze five different privacy-aware evaluation metrics and compare them for the three different experiments as described above. 

\subsubsection{Accuracy}
\begin{figure}[!t]
\centering
\includegraphics[width=\columnwidth]{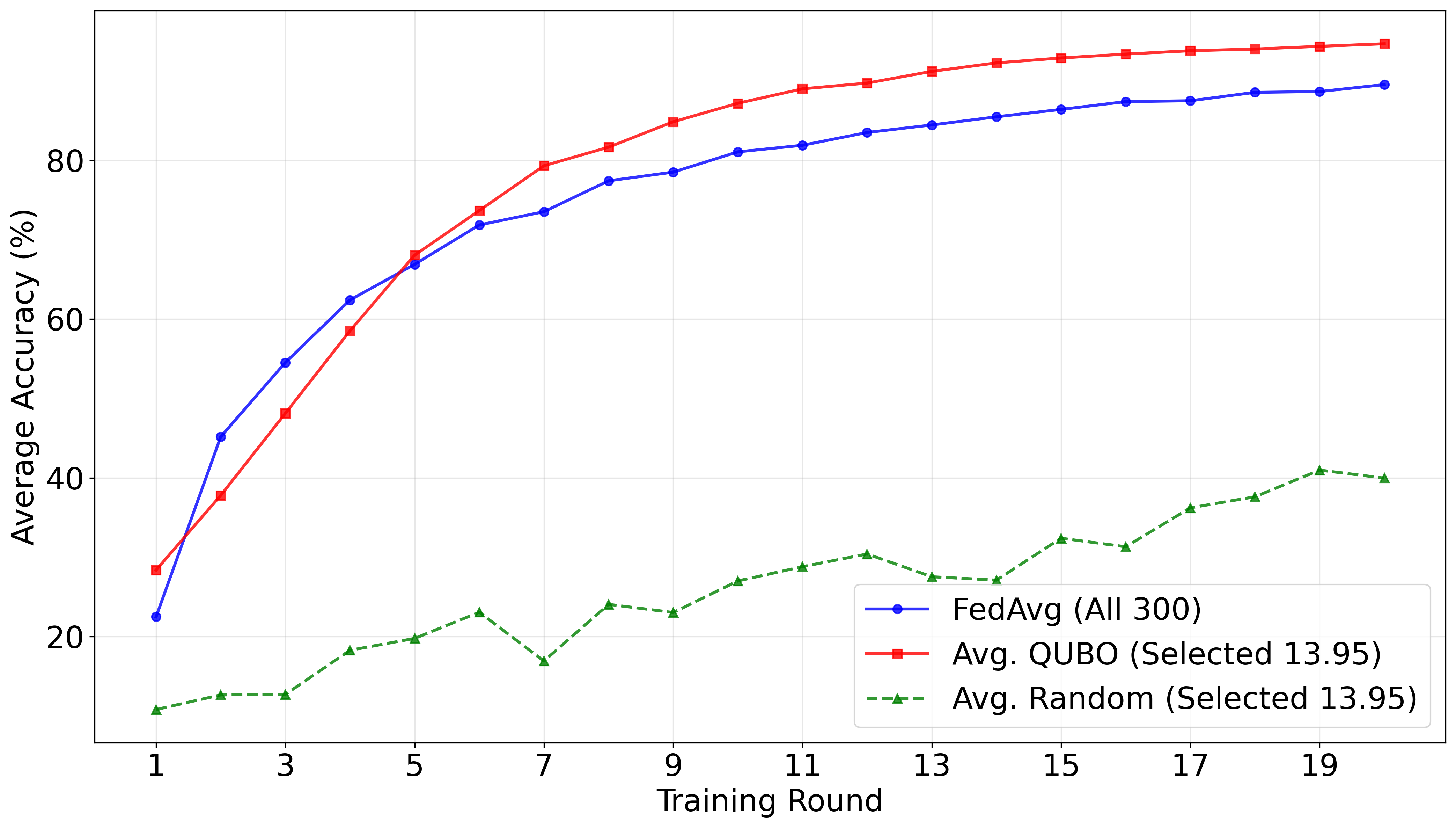}
\caption{Privacy-preserving training accuracy comparison between FedAvg, QUBO, and Random approaches across 20 rounds, averaged over 20 runs with different data heterogeneity levels ($\alpha$ values). QUBO maintains competitive accuracy while providing 95.2\% privacy preservation (in average, 13.95/300 clients expose gradients per round).}
\label{fig:privacy_accuracy_comparison}
\end{figure}

The accuracy benchmark comparison can be seen in Fig.~\ref{fig:privacy_accuracy_comparison}. This plot shows training accuracy evolution for 20 rounds with each round being performed over 20 runs. The QUBO selection strategy has better accuracy with only average 13.95 selected clients compared to FedAvg that uses all 300 clients. The QUBO accuracy surpass the Random selection by a large margin, mainly because of the extreme heterogeneity of the data distribution across clients. This demonstrates that the QUBO client selection method preserves privacy without compromising model quality.

\subsubsection{Loss}
\begin{figure}[!t]
\centering
\includegraphics[width=\columnwidth]{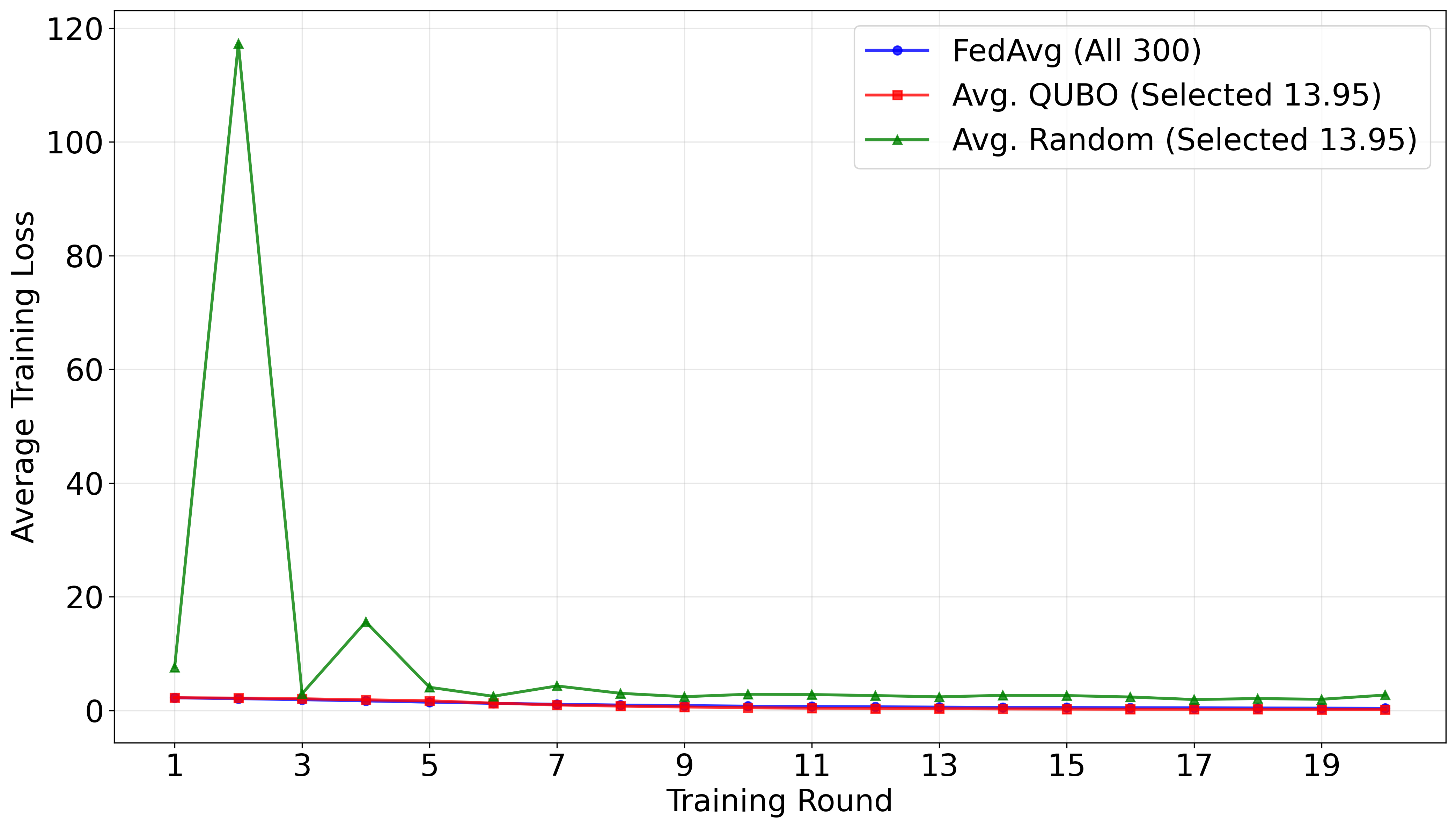}
\caption{Privacy-preserving training loss evolution for FedAvg and QUBO approaches over 20 rounds.}
\label{fig:privacy_loss_comparison}
\end{figure}

The loss evolution comparison across the training rounds is shown in Fig.~\ref{fig:privacy_loss_comparison}. The privacy-preserving QUBO selection method has better model convergence compared to the random-selection method and comparable to non-privacy-preserving FedAvg, making QUBO selection a superior choice.

\subsubsection{Data Heterogeneity}
\begin{figure}[!t]
\centering
\includegraphics[width=\columnwidth]{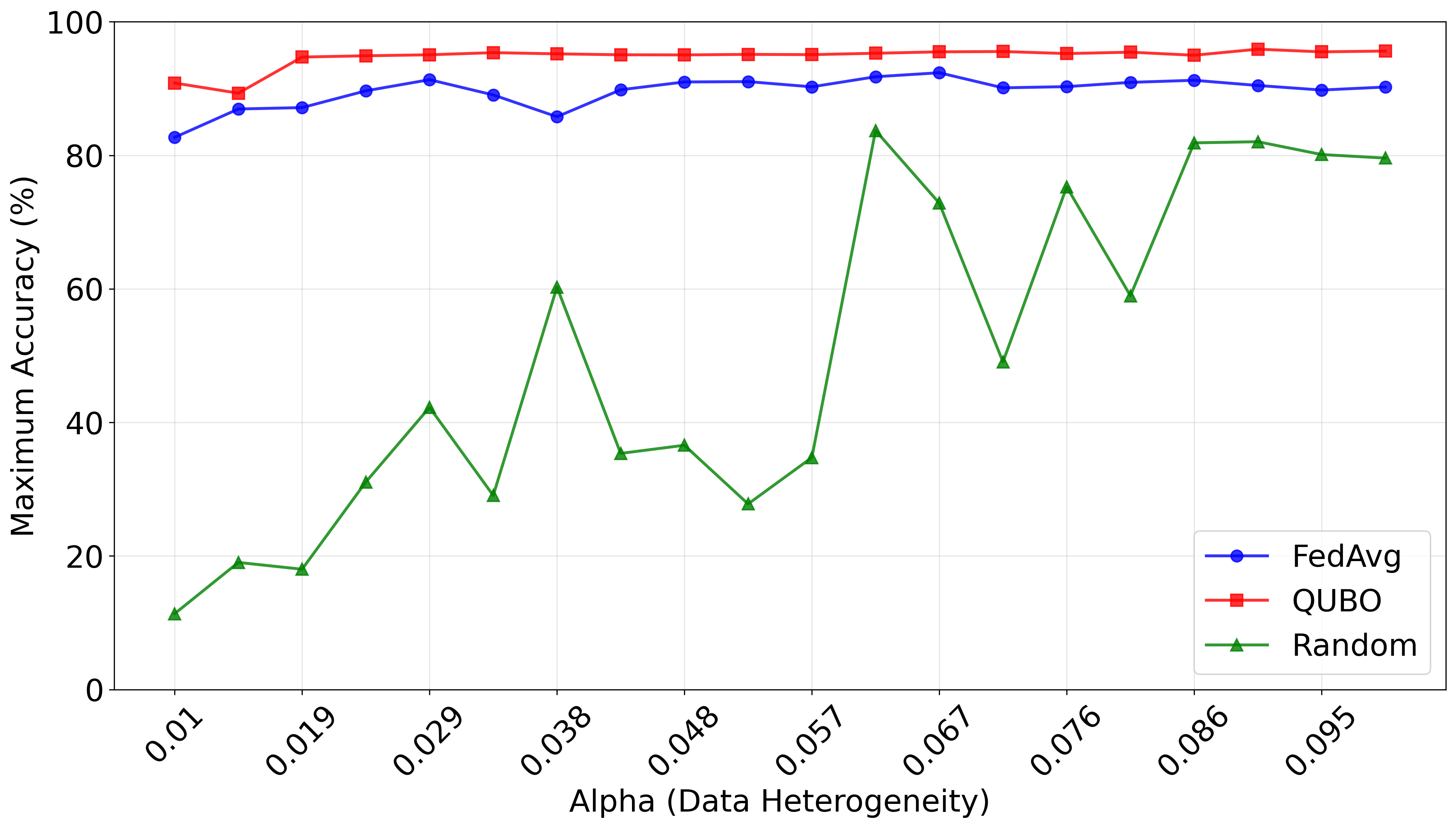}
\caption{MNIST maximum accuracy achieved under different data heterogeneity levels ($\alpha$ values) with privacy preservation. QUBO-based selection excels at all heterogeneity levels while maintaining 95.2\% per-round privacy preservation, demonstrating optimal privacy-utility tradeoffs even in challenging non-IID scenarios.}
\label{fig:privacy_heterogeneity_analysis}
\end{figure}

Fig.~\ref{fig:privacy_heterogeneity_analysis} represents model accuracy for several experiments with varying heterogeneity ($\alpha$). This plot revealed that QUBO method performance is not impacted by data heterogeneity.

\subsubsection{Gradient Quality}
\begin{figure}[!t]
\centering
\includegraphics[width=\columnwidth]{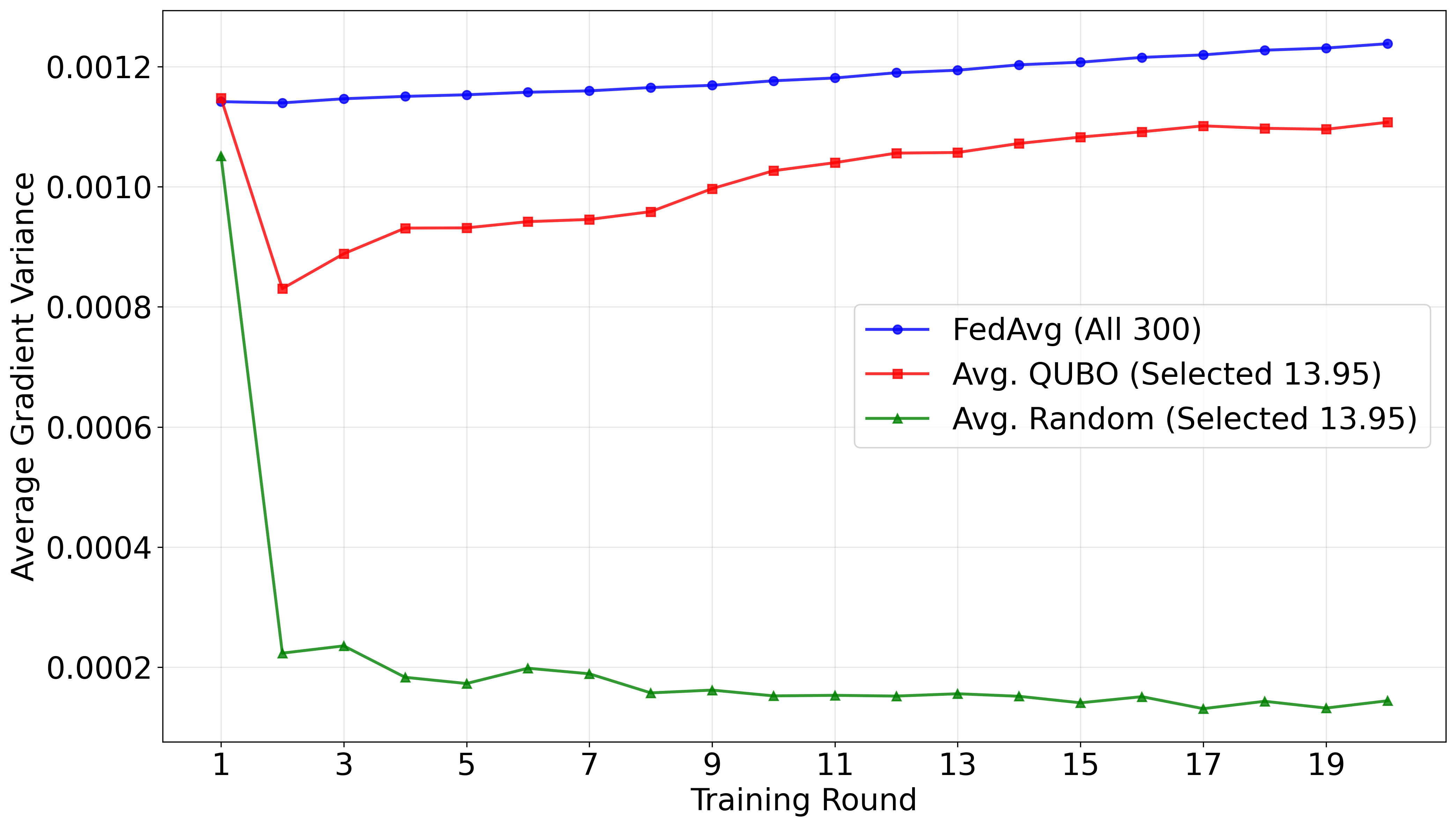}
\caption{MNIST gradient variance comparison under privacy constraints.}
\label{fig:privacy_gradient_variance}
\end{figure}

Fig.~\ref{fig:privacy_gradient_variance} shows that the full FedAvg has the highest variance, reflecting maximum diversity but also lack of stability. The QUBO method provides balance between stability and heterogeneity, whereas the Random method has the lowest variance, nevertheless, it is suboptimal for real FL tasks.

\subsubsection{Strategy Performance}
\begin{figure}[!t]
\centering
\includegraphics[width=\columnwidth]{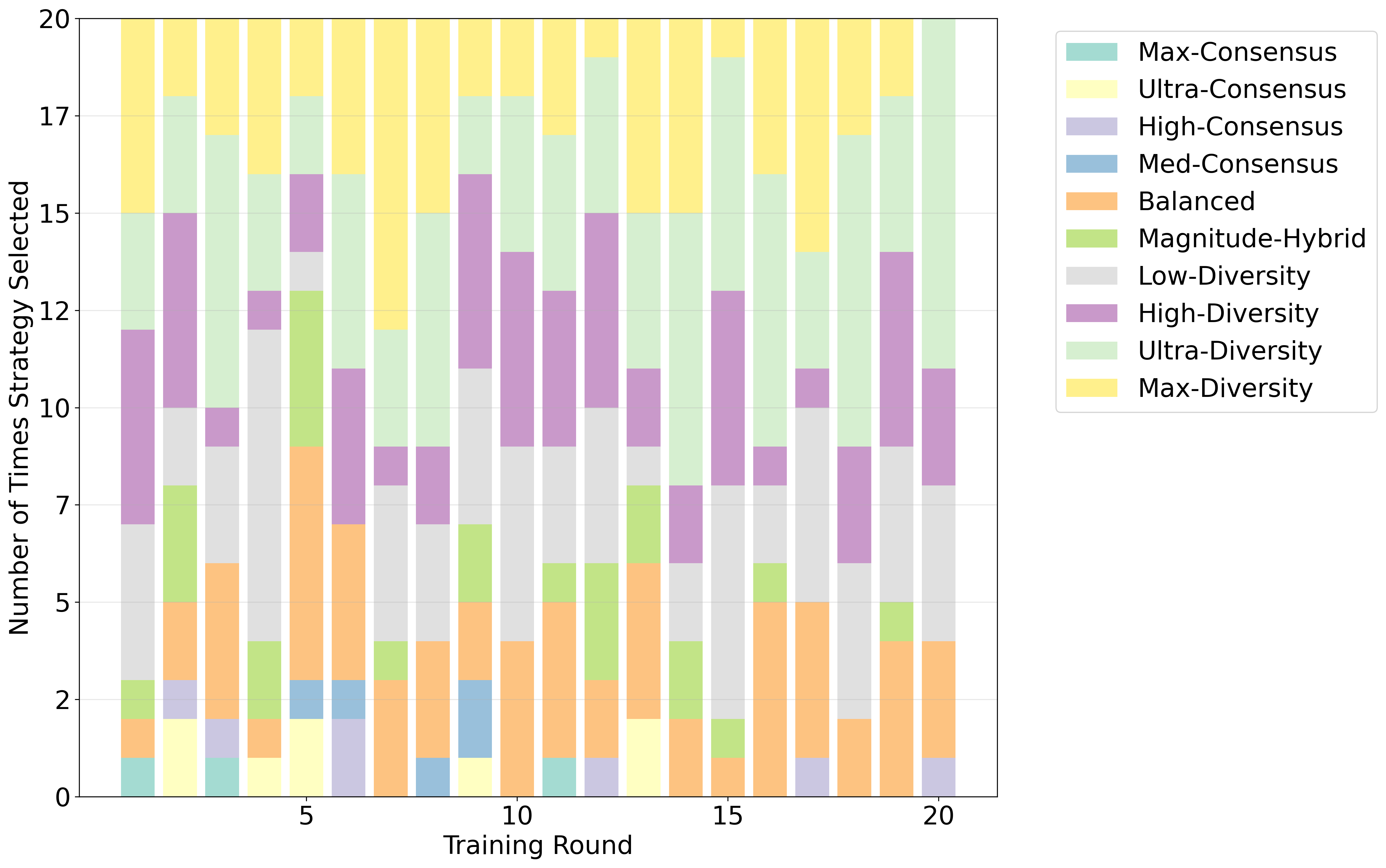}
\caption{MNIST QUBO strategy selection across training rounds. Our strategy selection scoring \ref{eq:scoring} favors exploration over exploitation for optimal privacy-utility balance.}
\label{fig:privacy_strategy_evolution}
\end{figure}

Fig.~\ref{fig:privacy_strategy_evolution} shows how QUBO strategies adapt across training rounds to optimize the privacy-utility tradeoff. Different strategies excel at different phases of training.

\begin{figure}[!t]
\centering
\includegraphics[width=\columnwidth]{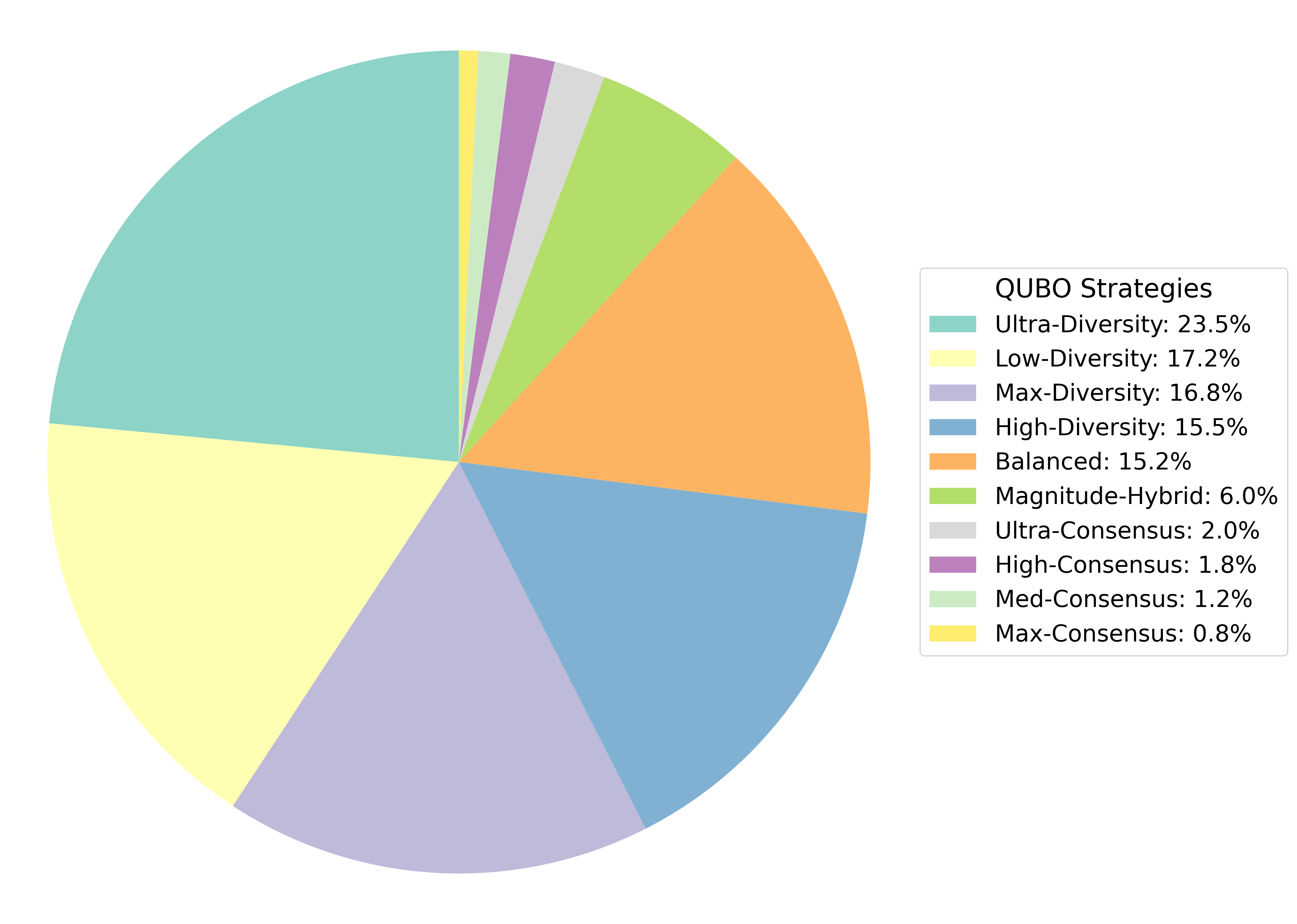}
\caption{MNIST distribution of winning privacy-preserving QUBO strategies on MNIST across all experimental conditions. Strategy selection (\ref{eq:scoring}) optimizes for both privacy preservation  and model performance.}
\label{fig:privacy_strategy_distribution}
\end{figure}

Fig.~\ref{fig:privacy_strategy_distribution} presents the overall distribution of privacy-preserving strategies across all experimental conditions. Our method extensively used all 10 strategies, a proof of the fact that all of them are useful. The distribution of strategies is dependent on factors, such as the model and the training data, and configuration needs to be based on such factors.

\subsection{Privacy Preservation Analysis}

\begin{figure}[!t]
\centering
\includegraphics[width=\columnwidth]{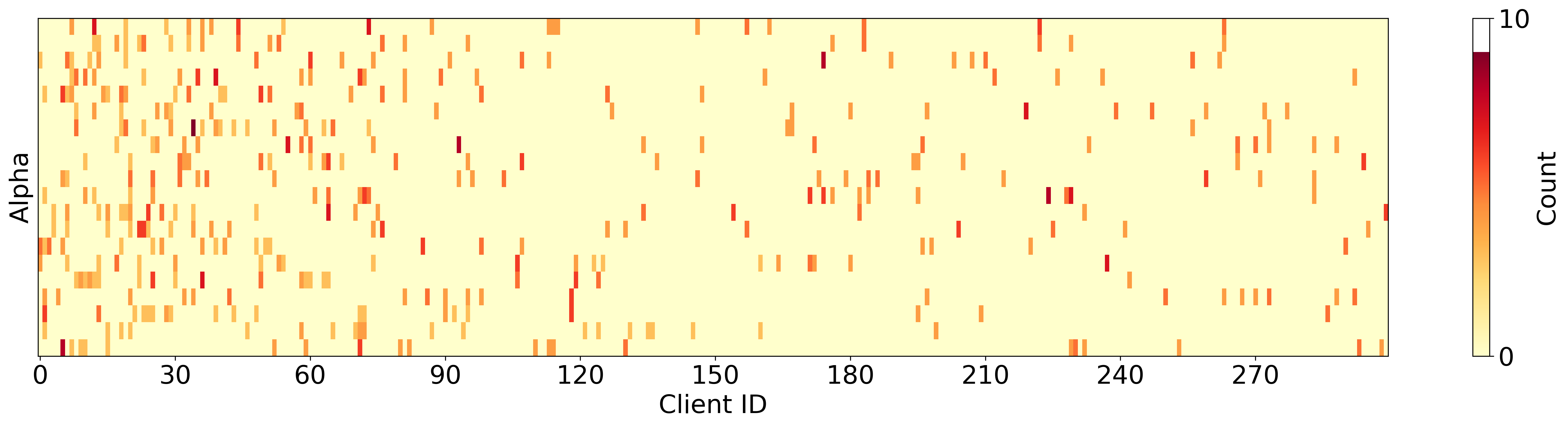}
\caption{Client selection heatmap for MNIST showing privacy preservation across 300 clients and different alpha values. The pattern demonstrates that ~153 clients (51\%) are ever selected across all rounds and heterogeneity levels, providing 95.2\% per-round privacy preservation while maintaining model performance.}
\label{fig:mnist_privacy_heatmap}
\end{figure}

\begin{figure}[!t]
\centering
\includegraphics[width=\columnwidth]{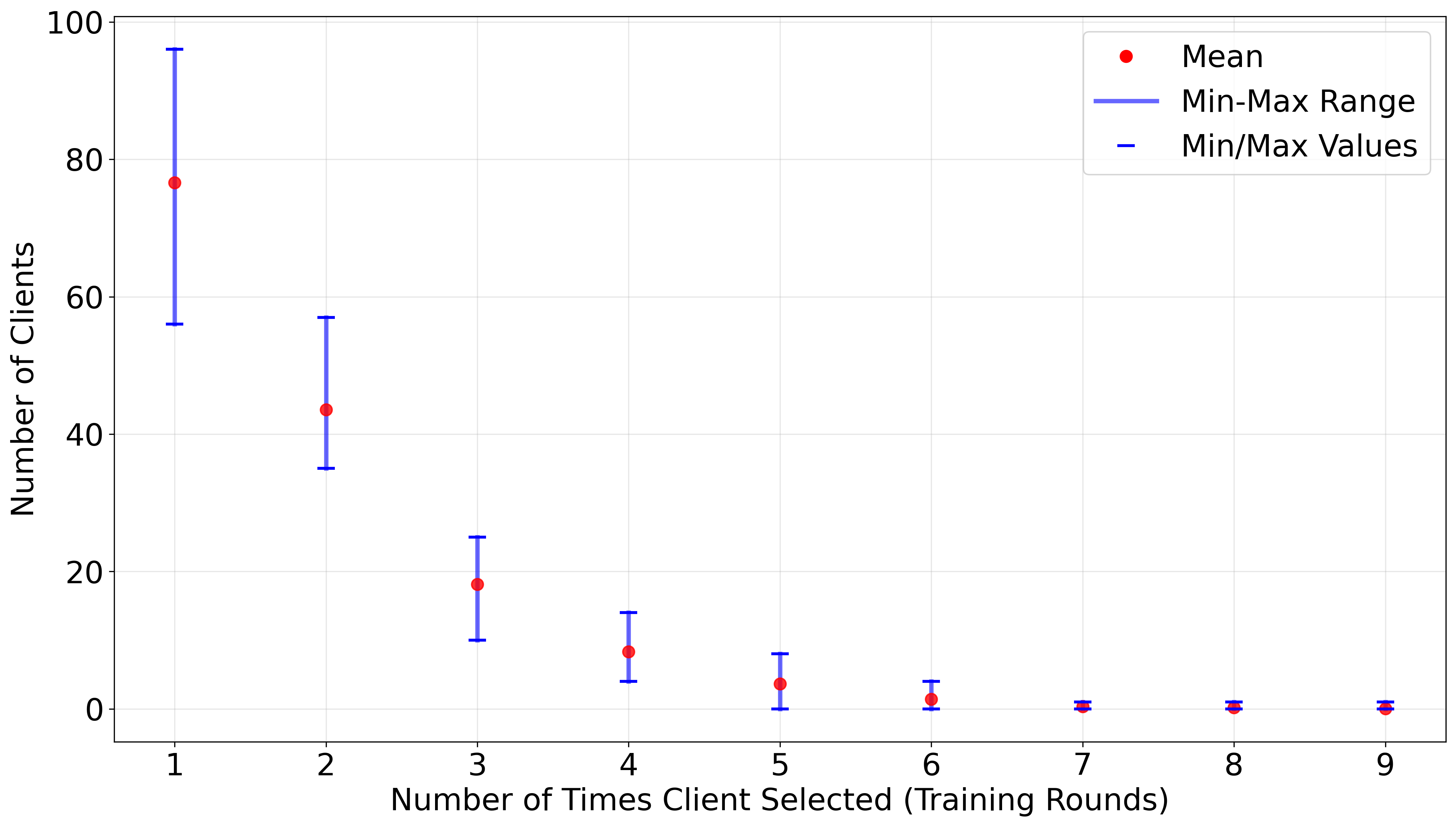}
\caption{Client frequency distribution for 20 runs with different $\alpha$ values for MNIST. Application logic \ref{eq:max-selections-exclusion} sets a hard limit on client exposure.}
\label{fig:client_frequency_analysis}
\end{figure}

Fig.~\ref{fig:mnist_privacy_heatmap} provides a visual representation of privacy preservation across all 300 MNIST clients and varying data heterogeneity levels. The sparse selection pattern clearly demonstrates that only a subset of clients are ever selected, with the rest maintaining complete privacy throughout training (49\% of all clients). The privacy benefits  achieved through strategic client selection in the MNIST experiment can be quantified as:
\begin{itemize}
\item Privacy preservation: 95.2\% of clients maintain complete privacy per round, meaning in average 13.95 clients reveal sensitive updates per round.
\item Cumulative privacy protection: Clients participate in only 5\% of rounds on average
\item Privacy-utility efficiency: Maintains model performance while minimizing privacy leakage
\item Adaptive privacy optimization: Strategy selection responds to data heterogeneity for optimal privacy-performance tradeoffs
\end{itemize}
We show the client selection frequency in Fig.~\ref{fig:client_frequency_analysis}. This plot shows that maximum number of client was selected only for single round of training. Number of selected clients decreases with increasing number of rounds. Even though we set the hard cut-off on a client selection at 10 rounds, only negligible number of clients were selected for more than six rounds of training, preserving privacy for the majority of clients as intended.

\subsection{CINIC-10 Experiments}
\begin{figure}[!t]
\centering
\includegraphics[width=\columnwidth]{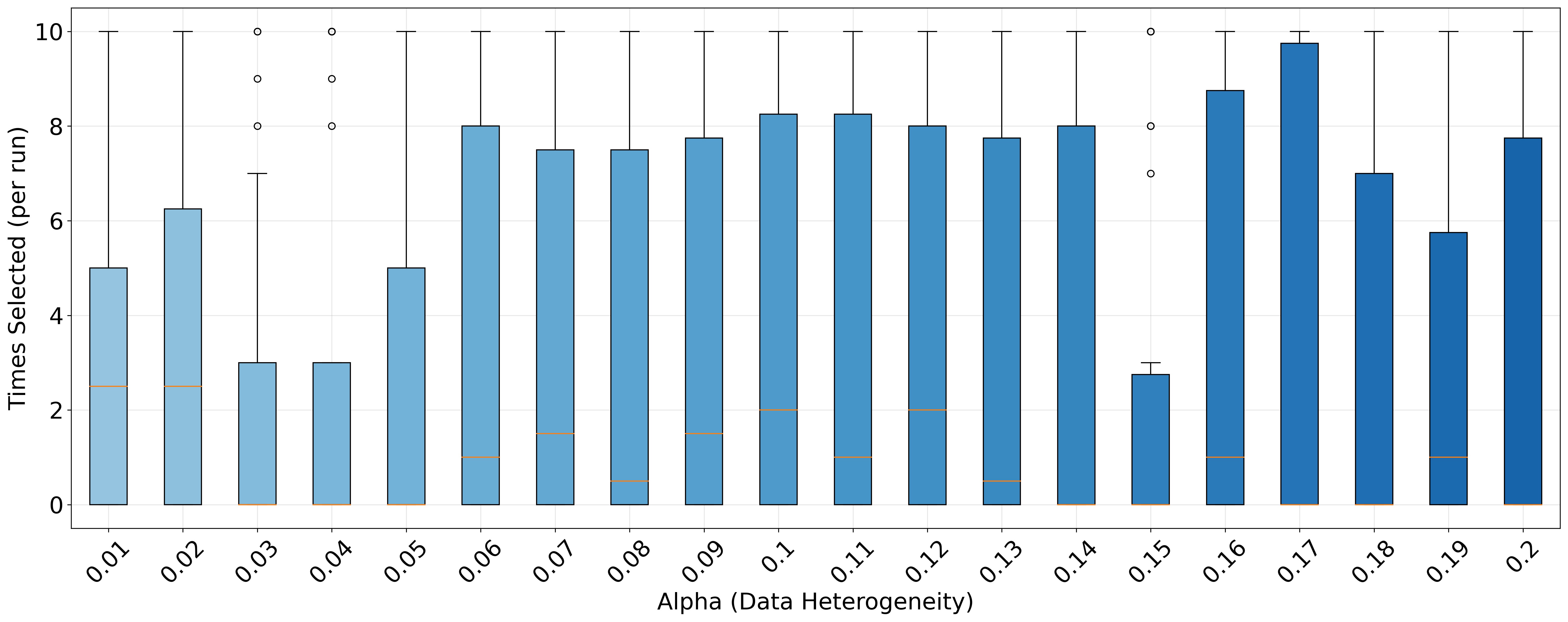}
\caption{Client fairness analysis for privacy-preserving selection on CINIC-10. Box plots show the distribution of client participation rates, demonstrating equitable privacy protection across all clients with QUBO-based selection achieving balanced privacy exposure  compared to random selection.}
\label{fig:cifar10_privacy_fairness}
\end{figure}

To validate our privacy-preserving approach across datasets, we conduct experiments on CINIC-10 \cite{darlow2018cinic10} with 30 total clients, QUBO selecting on average 5.5 clients per round for 33\% cumulative privacy preservation and 82\% per-round exposure reduction. We set a hard limit of 10 inclusions per client. Figure~\ref{fig:cifar10_privacy_fairness} demonstrates that our approach maintains fairness in privacy protection across clients even in more complex scenarios with fewer participants.

\section{Discussion}\label{sec:discussion}
% \subsection{Implications}
This paper introduces a novel QUBO-based approach to improve privacy in FL through strategic client selection. Our method formulates client selection as a QUBO problem that balances relevance, diversity, and redundancy across multiple distinct strategies. Experimental evaluation on MNIST with 300 clients demonstrates that our approach achieves 106\% of traditional FedAvg accuracy while reducing client participation by 95.2\% per round and 49\% cumulatively, thereby significantly reducing the risks of per-round and final model privacy exposure. The key contributions include: (1) formulation of FL client selection as a QUBO problem suitable for quantum optimization; (2) comprehensive strategy design spanning exploitation to exploration approaches; (3) demonstration of significantly improved privacy preservation with minimal model performance impact and even potential for improvement under various heterogeneity conditions.

Our QUBO client selection approach demonstrates that not all clients need to participate in the FL aggregation process to maintain model performance. This represents a novel approach for strengthening privacy of models generated via FL. While the results are encouraging, this method also suffers from various limitations. We list a few of them below:
\begin{enumerate}
    \item \textit{Scalability:} We demonstrate the experiments only on small dataset (300 clients for MNIST, and 30 clients for CINIC-10). We also use only simulated annealing for solving QUBO. Simulated annealing might be prohibitively costly for large-scale FL. Alternatively, we can use quantum annealer (QA) or gate-based quantum computer to solve QUBO. Usage of a quantum device might speed up the process, but the qubit number is limited on Noisy Intermediate-Scale Quantum (NISQ) devices.

    \item \textit{Server Trust Requirements:} While client selection reduces the number of exposed gradient updates to external observers, the server must still receive and process all client updates to make selection decisions. Therefore, privacy benefits are limited to scenarios where the server is trusted, and the approach does not provide privacy guarantees against server-based attacks or data reconstruction attempts by the aggregating server. Still, improved per-round privacy reduces the risk of honest but curious clients extracting insights from intermediary global model updates.

    \item \textit{No Formal Privacy Guarantee:} Method reduces but does not eliminate client exposure. Our experiments show, nevertheless, that 49\% of clients are fully protected this way. Selected clients' updates are aggregated into the model, potentially enabling reconstruction attacks on the global model itself. This approach is complementary to differential privacy techniques.
        
    \item \textit{Non-Byzantine Assumption:} We assume all clients are honest and do not consider adversarial or Byzantine participants. Extension to Byzantine-robust scenarios is left for future work.
\end{enumerate}
Our study was performed with a limited number of models and datasets; hence, there is no guarantee it will generalize to other circumstances or the fact that there are potentially better methods to achieve the same
results. In our experiments, the QUBO-based approach outperformed classical FedAvg, although results on two models and datasets with limited hyperparameter tuning are not sufficient to generalize. More experiments are needed and it is an effort we intend to pursue.

This work opens new directions for quantum-inspired privacy protection in distributed ML. As quantum hardware continues to advance, QUBO-based approaches may become increasingly practical for large-scale FL deployments.

\subsection*{Future Research Directions}
An easy extension would be to validate the results in this paper on actual quantum hardware as the hardware capabilities improve. Another interesting research direction would be to develop hierarchical or approximate QUBO formulations for large-scale federated networks. Such approximate QUBO formulations could also be used to fit a mid-scale network, like the ones we used in this work, to a NISQ device. We should also improve our QUBO formulations to handle adversarial clients so that our model is robust against Byzantine attacks. Both the datasets we used in our experiments are the image classification data. We can use similar idea on larger datasets and different ML tasks beyond image classification.

% \section{Conclusion}\label{sec:conclusion}
% \input{conclusion}

\bibliographystyle{IEEEtran}
\bibliography{references} % Changed to references.bib

\end{document}